%% file: squavi_iclr.tex
\documentclass{article} 
\usepackage{iclr2019_conference,times}

\input{math_commands.tex}

\usepackage{hyperref}
\usepackage{url}
\usepackage{graphicx}

\title{Predictive Uncertainty through Quantization}

\author{Bastiaan S. Veeling \\ University of Amsterdam \\ \texttt{basveeling@gmail.com} \And Rianne van den Berg \\ University of Amsterdam  \And Max Welling \\ University of Amsterdam 
}

\usepackage{amsmath,amssymb,bm}
\usepackage{booktabs}
\usepackage{wrapfig}
\usepackage[normalem]{ulem}
\renewcommand{\E}{\mathbb{E}}

\newcommand{\sv}{\mathbf{s}}
\newcommand{\xv}{\mathbf{x}}

\newcommand{\Dc}{\mathcal{D}}
\newcommand{\Lc}{\mathcal{L}}
\newcommand{\zv}{\mathbf{z}}
\newcommand{\yv}{\mathbf{y}}
\renewcommand{\vv}{\mathbf{v}}
\newcommand{\Wv}{\mathbf{W}}
\newcommand{\wv}{\mathbf{w}}
\newcommand{\bv}{\mathbf{b}}
\DeclareMathOperator*{\gumbelsoft}{gumb}
\DeclareMathOperator*{\soft}{softmax}

\newcommand{\thetav}{\mathbf{\theta}}

\begin{document}

\maketitle

\begin{abstract}
High-risk domains require reliable confidence estimates from predictive models. 
Deep latent variable models provide these, but suffer from the rigid variational distributions used for tractable inference, which err on the side of overconfidence.
We propose Stochastic Quantized Activation Distributions (SQUAD), which imposes a flexible yet tractable distribution over discretized latent variables.
The proposed method is scalable, self-normalizing and sample efficient. We demonstrate that the model fully utilizes the flexible distribution, learns interesting non-linearities, and provides predictive uncertainty of competitive quality.

\end{abstract}

\section{Introduction}
\begin{wrapfigure}{r}{0.25\textwidth}
\vspace{-2.2\baselineskip}
\includegraphics[width=0.25\textwidth]{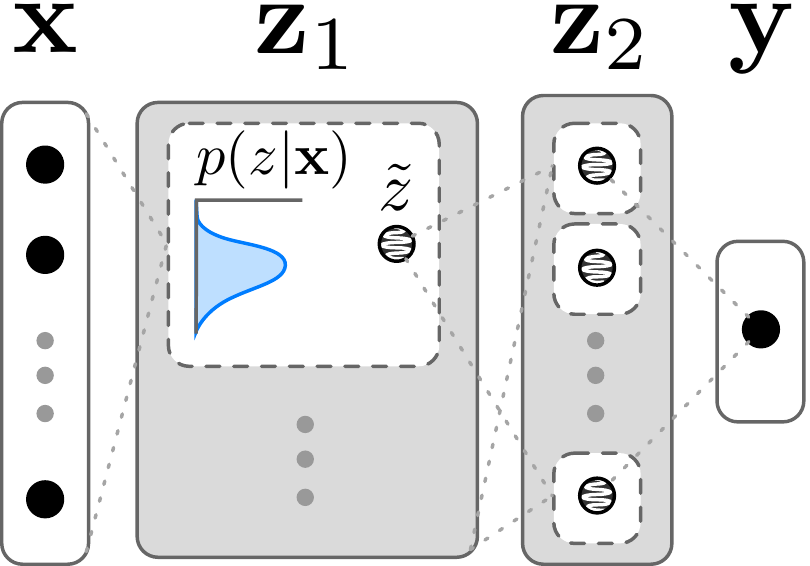}
\vspace{-1.5\baselineskip}\small
\caption{DLVMs have layers of stochastic latent variables.}
\label{fig:intuition_dlvm}
\vspace{-1.\baselineskip}
\end{wrapfigure}
In high-risk domains, prediction errors come at high costs. Luckily such domains often provide a failsafe: self-driving cars perform an emergency stop, doctors run another diagnostic test, and industrial processes are temporarily halted. For deep learning models, this can be achieved by rejecting datapoints with a confidence score below a predetermined threshold. This way, a low error rate can be guaranteed at the cost of rejecting some predictions. However, estimating high quality confidence scores from neural networks, which create well-ordered rankings of correct and incorrect predictions, remains an active area of research.

Deep Latent Variable Models (DLVMs, fig.~\ref{fig:intuition_dlvm}) approach this by postulating latent variables $\zv$ for which the uncertainty in $p(
\zv|\xv)$ influences the confidence in the target prediction. 
Recently, efficient inference algorithms have been proposed in the form of variational inference,  where an inference neural network is optimized to predict parameters of a variational distribution that approximates an otherwise intractable distribution (\cite{Kingma2013-fy, Rezende2014-in, Alemi2016-hp, Achille2016-tm}).

\begin{wrapfigure}[11]{l}{0.2\textwidth}
\vspace{-1\baselineskip}
\includegraphics[width=0.2\textwidth]{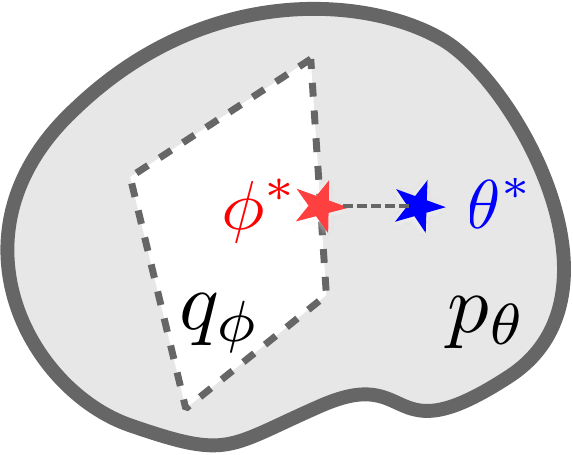}
\vspace{-1.5\baselineskip}\small
\caption{ A distribution $q_\phi$ is optimized to approximate $p_{\theta^*}$.}
\label{fig:intuition_var}
\vspace{-1.\baselineskip}
\end{wrapfigure}
\begin{wrapfigure}{r}{0.35\textwidth}
\vspace{-1.3\baselineskip}
\includegraphics[width=0.35\textwidth]{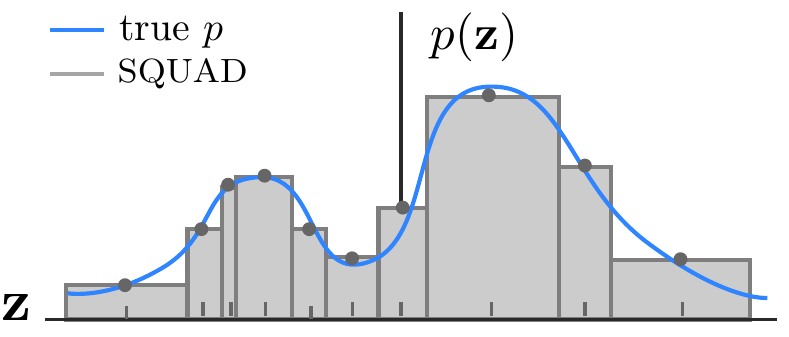}
\vspace{-1.5\baselineskip}\small 
\caption{SQUAD quantizes the domain of $z$ to model a flexible and tractable variational distribution.}
\label{fig:intuition_squad}
\end{wrapfigure}

Variational inference relies on a tractable class of distributions that can be optimized to closely resemble the true distribution (fig.~\ref{fig:intuition_var}), and it's hypothesized that more flexible classes lead to more faithful approximations and thus better performance (\cite{Jordan1999-et}). To explore this hypothesis, we propose a novel tractable class of highly flexible variational distributions. Considering that neural networks with low-precision activations exhibit good performance (\cite{Holi1993-gq,Hubara2016-qr}), we make the modeling assumption that latent variables can be expressed under a strong quantization scheme, without loss of predictive fidelity. If this assumption holds, it becomes tractable to model a scalar latent variable with a flexible multinomial distribution over the quantization bins (fig.~\ref{fig:intuition_squad}).

By re-positioning the variational distribution from a potentially limited description of moments, as found in commonly applied conjugate distributions, to a direct expression of probabilities per value, a variety of benefits arise. As the output domain is constrained, the method becomes self-normalizing, relieving the model from hard-to-parallelize batch normalization techniques (\cite{Ioffe2015-ib}). More interesting priors can be explored and the model is able to learn unique activation functions per neuron. 

More concretely, the contributions of this work are as follows:

\begin{itemize}
\item We propose a novel variational inference method by leveraging multinomial distributions on quantized latent variables.
\item We show that the emerging predicted distributions are multimodal, motivating the need for flexible distributions in variational inference.
\item We demonstrate that the proposed method applied to the information bottleneck objective computes competitive uncertainty over the predictions and that this manifests in better performance under strong risk guarantees.
\end{itemize}

\section{Background}
\label{sec:back}

In this work, we explore deep neural networks for regression and classification. We have data-points consisting of intputs $\xv$ and targets $\yv$ in a dataset $\Dc = \{(\xv^i, \yv^i)~|~i \in [1, ...,  N]\}$ and postulate latent variables $\zv$ that represent the data. We focus on the Information Bottleneck (IB) perspective: first proposed by \cite{Tishby2000-rk}, the information bottleneck objective $I(\yv,\zv; \theta) - \beta I(\xv,\zv; \theta)$ is optimized to maximize the mutual information between $\zv$ and $\yv$, whilst minimizing the mutual information between $\zv$ and $\xv$. The objective can be efficiently optimized using a variational inference scheme as shown concurrently by both \cite{Alemi2016-hp} and \cite{Achille2016-tm}. Under the Markov assumption $P(\zv, \xv, \yv) = p(\zv|\xv)p(\yv|\xv)p(\xv)$, they derive the following lower bound: 
\begin{align}
& I(\yv,\zv; \theta) - \beta I(\xv,\zv; \theta)
\geq \Lc = \frac{1}{N} \sum_{n=1}^N \E_{p_\theta(\zv|\xv_n)}[\log q_\theta(\yv_n | \zv)] - \beta\KL [p_\theta(\zv|\xv_n) \| r(\zv) ] \label{eq:elbo},
\end{align}
where $\E_{p(\zv|\xv_n)}$ is commonly estimated using a single Monte Carlo sample and $r(\zv)$ is a variational approximation to the marginal distribution of $\zv$. In practice $r(\zv)$ is fixed to a simple distribution such as a spherical Gaussian. 
\cite{Alemi2016-hp} and \cite{Achille2016-tm} continue to show that the Variational Auto Encoder (VAE) Evidence Lower Bound (ELBO) proposed in \cite{Kingma2014-ll, Rezende2014-in} is a special case of the IB bound when $\yv = \xv$ and $\beta = 1$:
\begin{align}
I(z,x) - \beta I(z,i) \geq \E_{p_\theta(\zv|\xv_n)}[\log q_\thetav(\xv_n | \zv)] - \KL [p_\theta(\zv|\xv_n) \| r(\zv) ] \label{eq:vaeelbo},
\end{align}
where $i$ represents the identity of data-point $\xv_i$.
Interestingly, the VAE perspective considers the bound to optimize a variational distribution $q(\zv|\xv)$, whilst the IB perspective prescribes that $q(\zv|\xv)$ in the ELBO is not a variational posterior but the true encoder $p(\zv|\xv)$, and instead $p(\yv|\zv)$ and $p(\zv)$ are the distributions approximated with variational counterparts. 

From yet another perspective, equation~\ref{eq:elbo} can be interpreted as a domain-translating beta-VAE (\cite{Higgins2016-vn}), where an input image is encoded into a latent space and decoded into the target domain. The Lagrange multiplier $\beta$ then controls the trade-off between rate and distortion, as argued by \cite{Alemi2017-ts}.

In this work, we follow the IB interpretation of the bound in equation~\ref{eq:elbo} and leave the evaluation of our proposed variational inference scheme in other models such as the VAE for further work.

\section{Method}
\label{sec:method}
\begin{figure}
\begin{center}
\centerline{\includegraphics[width=1.0\textwidth]{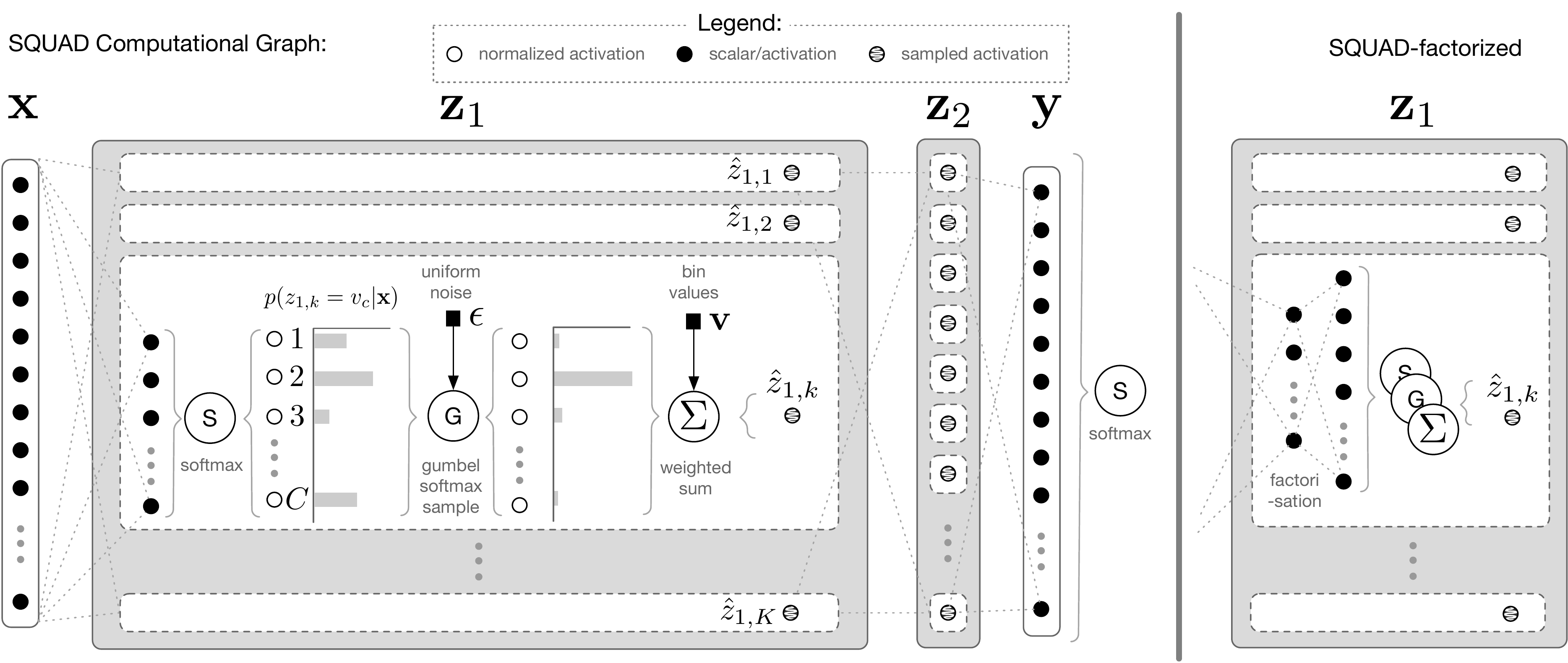}}
\caption{The left diagram visualizes the computational graph of SQUAD at training time, providing a detailed view on how an individual latent variable is sampled. The right diagram visualizes how the proposed matrix-factorization variant improves the parameter efficiency of the model.}
\label{fig:diagram}
\end{center}
\end{figure}
At the heart of our proposal lies the assumption that neuron networks can be effective under strong activation quantization schemes. We start with presenting the derivation if the model in the context of a single latent-layer information bottleneck, following the single data-point loss in \eqref{eq:elbo} and dropping the subscript $_n$ for clarity, with figure~\ref{fig:diagram} for visual reference:
\begin{align} 
\Lc=& \E_{p_\theta(\zv|\xv)}[\log q_\theta(\yv | \zv)] - \beta\KL [p_\theta(\zv|\xv) \| r(\zv) ] . \label{eq:ourbo}
\end{align}
 To impose a flexible, multi-modal distribution over $\zv$, we first make a mean-field assumption $p(\zv | \xv) = \prod_k p(z_k | \xv)$. We then quantize the \textit{domain} of each of the $K$ scalar latent variables \(z_k\) such that only a small set of potential values  remain: \(z_k \in \vv = \{v_1, \dots, v_C\}~~ \forall k\) with e.g.\ \(C=11\), see fig.~\ref{fig:intuition_squad}.  

To optimize the parameters $\thetav$ with Stochastic Gradient Descent (SGD), we need to derive a fully differentiable sampling scheme that allows us to sample values of $\zv$. To formulate this, we re-parametrize the expectation over $\zv$ in \eqref{eq:ourbo} using a set of variables $s_k \in \lbrace 1, \dots,  C\rbrace$ which index the value vector $\vv$, allowing us to use a softmax function to represent the distribution over $\zv$:
\begin{align} 
p_\thetav(z_k = v_c | \xv) \triangleq  p_\thetav(s_k = c | \xv) = \soft(\Wv\xv + \bv)_c. \label{eq:sdef}
\end{align}
These indexing values $\sv$ are then used in conjunction with values $\vv$ as in input for $q_\thetav(\yv|\zv)$, which is modelled with a small network $f_\thetav(\cdot)$:  
(\textit{abusing notation to indicate element-wise indexing with }$\vv[\sv]$):
\begin{align} 
\Lc = \E_{\sv \sim p_\thetav}[\log f_\theta\left(\vv[\sv ] \right) ] - \beta\KL [\ldots] \nonumber.
\end{align}
To enable sampling from the discrete variables $\sv$, we use the Gumbel-Max trick (\cite{Gumbel1954-pf}), denoted $\gumbelsoft()$, re-parameterizing the expectation $\E_{\sv \sim p}$ with uniform noise $\epsilon \sim U(0,1)$:
\begin{align} 
= \E_{\epsilon} \left[\log f_\theta\left( \vv\left[ \argmax_c \gumbelsoft(p_\thetav(\sv| \xv), \epsilon)\right]\right)\right] - \beta\KL [\ldots].
\end{align}
As the argmax is not differentiable, we approximate this expectation using the Gumbel-Softmax trick (\cite{Maddison2016-hn,Jang2016-ek}), which generates samples that smoothly deform into one-hot samples as the softmax temperature $\tau$ approaches $0$. Using the inner product (denoted $\cdot$) of the approximate one-hot samples and $\vv$, we create samples from $\zv$:
\begin{align} 
\approx \E_{\epsilon} \left[\log f_\theta\left( \vv \cdot \soft_c  \gumbelsoft\large(p_\thetav(\sv | \xv), \epsilon\large)\right)\right] - \beta\KL [\ldots]. \label{eq:ourelboq}
\end{align}
In practice, we anneal $\tau$ from $1.0$ to $0.5$ during the training process, as proposed by \cite{Yang2017-hw} to reduce gradient variance initially, at the risk of introducing bias.

\begin{figure}
\begin{center}
\centerline{\includegraphics[width=1\textwidth]{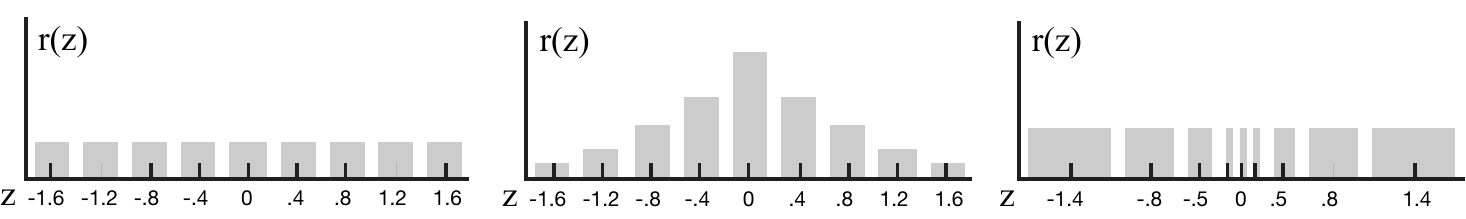}}
\caption{
In the IB bound, the marginal $p(z)$ is approximated with a fixed distribution $r(z)$. Using our proposed SQUAD distribution we can impose a variety of interesting forms for $r(\zv)$ via the spacing $\vv$ and weighting $r(z_k = v_c)$ of the quantization bins. For the values $\vv$, we compare linearly spaced bins \textbf{(left)} versus bins with equal probability mass under a normal distribution \textbf{(right)}. Furthermore, we explore the effect of allowing the bin values to be optimized with SGD on a per-neuron or per-layer basis, to allow the model to optimize the quantization scheme with the highest fidelity. For the prior probabilities, we explore a uniform prior \textbf{(left)} and probability mass of the bins under a normal distribution \textbf{(middle)}.
\label{fig:priors}}
\end{center}
\end{figure}

To conclude our derivation, we use a fixed SQUAD distribution to model the variational marginal $r(\zv)$ as shown in figure~\ref{fig:priors}. We can then derive the KL term analytically following the definition for discrete distributions. Using the fact that the KL divergence is additive for independent variables, we get our final loss:
\begin{align} 
\Lc=&\E_{\epsilon} \left[\log f_\theta\left( \vv \cdot \soft_c  \gumbelsoft\large(p_\thetav(\sv | \xv), \epsilon\large)\right)\right] -\beta\sum_{k=1}^K \sum_{c=1}^C p_\thetav(s_k = c | \xv) \log \frac{p_\thetav(s_k = c | \xv)}{r(z_k = v_c)}. \label{eq:ourelbokl}
\end{align}
For the remainder of this work, we will refer to the latent variables as $\zv$ in lieu of $\sv$, for clarity. 

At test time, we can approximate the predictive function $p(y^* | \xv^*)$ for a new data-point $\xv^*$ by taking $T$ samples from the latent variables $\zv$  i.e.\ $\hat{\zv}_t \sim p(\zv | \xv^*)$, and averaging the predictions for $y^*$:
\begin{align}
p_\thetav(y^* | \xv^*) &\approx \int q_\thetav(y^* | \zv) p_\thetav(\zv | \xv^*) d\zv 
\approx \dfrac{1}{T}  \sum^{T}_{t=1} q_\thetav(y^* | \hat{\zv}_t).
\end{align}
We improve the flexibility of the proposed model by creating a hierarchical set of latent variables. The joint distribution of L layers of latents is then:
\begin{align}
p_\thetav(\zv_1, \dots, \zv_L | \xv) &=p_{\theta}(\zv_L|\zv_{L-1}) \cdots p_{\theta}(\zv_1|\xv), 
\end{align}
With $q_\thetav(\yv | \zv_1, \dots, \zv_L) =q_\thetav(\yv | \zv_L)$. This is straightforwardly implemented with a simple ancestral sampling scheme. 

Interestingly, the strong quantization proposed in our method can itself be considered an additional information bottleneck, as it exactly upper-bounds the number of bits per latent variable. Such bottlenecks are theorized to have a beneficial effect on generalization (\cite{Tishby2000-rk, Achille2016-tm, Alemi2017-ts, Alemi2016-hp}), and we can directly control this bottleneck by varying the number of quantization bins. 

The computational complexity of the method, as well as the number of model parameters $\theta$, scale linearly in $C$, i.e. $O(C)$ (with $C$ the number of quantization bins). It is thus suitable for large-scale inference. We would like to stress that the proposed method differs from work that leverages the Gumbel-Softmax trick to model categorical latent variables: our proposal models continuous scalar latent variables by quantizing their domain and modeling belief with a multinomial distribution. Categorical latent variable models would incur a much larger polynomial complexity penalty of $O(C^2)$.

\paragraph{Matrix-factorization variant}
To improving the tractability of using a large number of quantization bins, we propose a variant of SQUAD that uses a matrix factorization scheme to improve the parameter efficiency. Formally, \eqref{eq:sdef} becomes:
\begin{align}
p(s_k = c | \xv) = \soft( \wv^{\prime\prime}_{k,c} (\wv^\prime_k\xv + \bv^\prime_k)  + \bv^{\prime\prime}_{k,c}) \nonumber,
\end{align}
with full layer weights $\Wv^{\prime\prime}$ and $\Wv^\prime$ respectively of shape $(K, B, C)$ and $(|X|, K, B)$, where $K$ denotes the number of neurons, $B$ the number of factorization inputs, $C$ number of quantization bins and $|X|$ the input dimensionality.
To improve the parameter efficiency, we can learn $\Wv^\prime$ per layer as well, resulting in shape $(|X|, 1, B)$, which is found to be beneficial for large C by the hyper-parameter search presented in section~\ref{sec:results}.
We depict this alternative model on the right side of figure~\ref{fig:diagram} and will refer to it as SQUAD-factorized. We leave further extensions such as Network-in-Network (\cite{Lin2013-kd}) for future work.

\section{Related Work}
Outside the realm of DLVMs, other methods have been explored for predictive uncertainty. \citet{Lakshminarayanan2017-pt} propose deep ensembles: straightforward averaging of predictions from a small set of separately adversarially trained DNNs. Although highly scalable, this method requires retraining a model up to 10 times, which can be inhibitively expensive for large datasets. 

\citet{Gal2015-th} propose the use of dropout (\cite{Srivastava2014-am}) at test time and present a Bayesian neural network interpretation of this method. A follow-up work by \citet{Gal2017-oh} explores the use of Gumbel-Softmax to smoothly deform the dropout noise to allow optimization of the dropout rate during training. A downside of MC-dropout is the limited flexibility of the fixed bi-modal delta-peak distribution imposed on the weights, which requires a large number of samples for good estimates of uncertainty.
\citet{Van_den_Oord2017-fz} propose the use of vector quantization in variational inference, quantizing a multi-dimensional embedding, rather than individual latent variables, and explore this in the context of auto-encoders.

In the space of learning non-linearity's, \cite{Su2017-mm} explore a flexible non-linearity that can assume the form of most canonical activations. 
More flexible distributions have been explored for distributional reinforcement learning by \cite{Dabney2017-qu} using quantile regression, of which can be seen as a special case of SQUAD where the bin values are learned but have fixed uniform probability. 
Categorical distributions on scalar variables have been used to model more flexible Bayesian neural network posteriors as by \cite{Shayer2017-vr}. The use of a mixture of diracs distribution to approximate a variety of distributions was proposed by \cite{Schrempf2006-re}.

\section{Results}
\label{sec:results}
\begin{figure}
\begin{center}
\centerline{\includegraphics[width=.8\columnwidth]{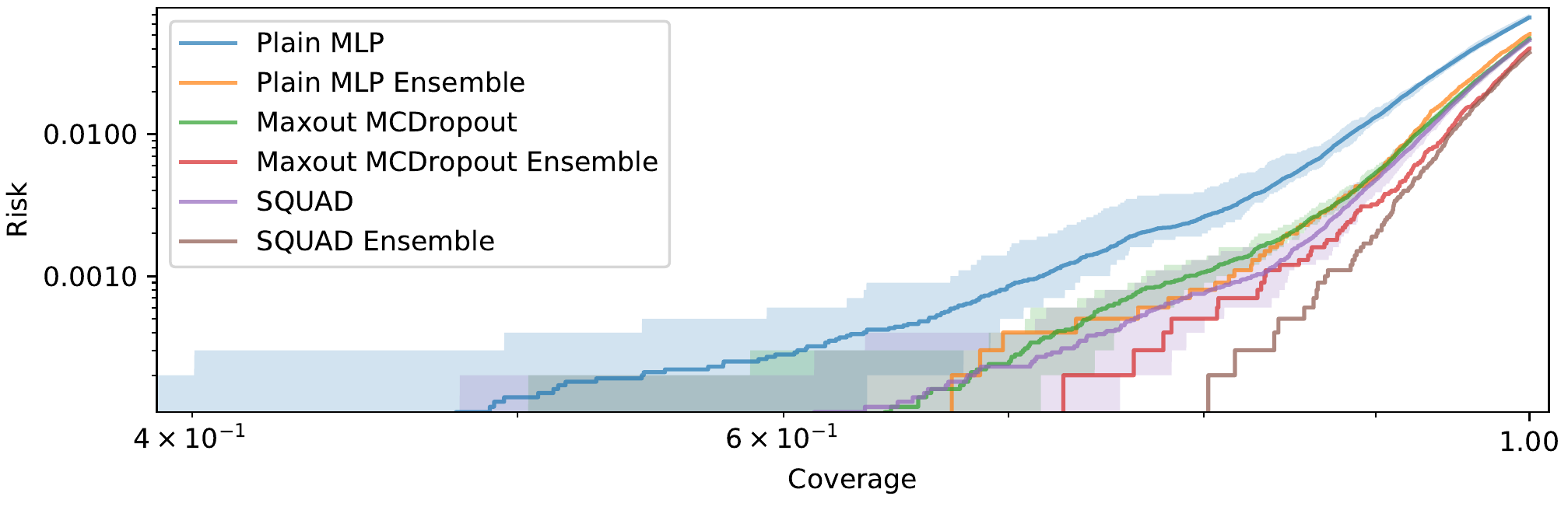}}
\caption{Risk/coverage curve (with log-axes for discernibility) of 2-layer models on notMNIST. Lines closer to the lower-right are better. As the selective classifier lowers the confidence threshold, coverage increase at the cost of greater classification risk. Area around curves represents 90\% confidence bounds computed using 10 initializations/splits.} 
\label{fig:riskcov1}
\end{center}
\end{figure}

Quantifying the quality of uncertainty estimates of models remains an open problem. Various methods have been explored in previous works, such as relative entropy \cite{Louizos2017-dw, Gal2015-ju}, probability calibration, and proper scoring rules \cite{Lakshminarayanan2017-pt}. Although interesting in their own right, these metrics do not directly measure a good ranking of predictions, nor indicate applicability in high-risk domains. Proper scoring rules are the exception, but a model with good ranking ability does not necessarily exhibit good performance on proper scoring rules: any score that provides relative ordering suffices and does not have to reflect true calibrated probabilities. In fact, well-ranked confidence scores can be re-calibrated (\cite{Niculescu-Mizil2005-ap}) after training to improve performance on proper scoring rules and calibration metrics.

In order to evaluate the applicability of the model in high-risk fields such as medicine, we want to quantify how models perform under a desired risk requirement. We propose to use the \textit{selection with guaranteed risk} (SGR) method\footnote{We deviate slightly from \cite{Geifman2017-hp} in that we use Softmax Response (SR) -- the probability taken from the softmax output for the most likely class -- as the confidence score for all methods. \cite{Geifman2017-hp} instead proposed to use the variance of the probabilities for MCdropout, but our experiments showed that SR paints MCdropout in a more favorable light.} introduced by \citet{Geifman2017-hp} to measure this. In summary, the SGR method defines a selective classifier using a trained neural network and can guarantee a user-selected desired risk with high probability (e.g.\ 99\%), by selectively rejecting data points with a predicted confidence score below an optimal threshold. 

To limit the influence of hyper-parameters on the comparison, we use the automated optimization method TPE (\cite{Bergstra2011-ri}) for both baselines and our models. The hyper-parameters are optimized for coverage at 2\% risk ($\delta=0.01$) on fashionMNIST, and sub-sequentially evaluated on notMNIST. Larger models are evaluated on SVHN.

We compare\footnote{All models are optimized with ADAM (\cite{Kingma2014-ll}), weight initialization as proposed by \cite{He2015-hz}, a weight decay of $10^{-5}$ and adaptive learning rate decay scheme | 10x reduction after 10 epochs of no validation accuracy improvement| and use early stopping after 20 epochs of no improvement.} our model against plain MLPs, MCDropout using Maxout activations\footnote{We found this baseline to perform stronger in comparison to conventional ReLU MCdropout models, under equal number of latent variables.} (\cite{Goodfellow2013-qw,Chang2015-mi}) and an information bottleneck model using mean-field Gaussian distributions. We evaluate the complementary deep ensembles technique (\cite{Lakshminarayanan2017-pt}) for all methods.

\subsection{Main results}
We start our analysis by comparing the predictive uncertainty of 2-layer models with 32 latent variables per layer. In figure~\ref{fig:riskcov1} we visualize the risk/coverage trade-off achieved using the predicted uncertainty as a selective threshold, and present coverage results in table~\ref{tab:cov}. Overall, we find that SQUAD performs significantly better than plain MLPs and deep Gaussian IB models, and we tentatively attribute this to the increased flexibility of the multinomial distribution. Compared to a Maxout MCdropout model with a similar number of weights, SQUAD appears to have a slight |though not significant| advantage, despite the strong quantization scheme, especially at low risk. Deep ensembles improve results for all methods, which fits the hypothesis that ensembles integrate over a form of weight uncertainty. When evaluated on a new dataset without retuning hyperparameters, SQUAD shows strong performance, as shown in table~\ref{tab:cov_notmnist}.

\begin{table}
\caption{SGR coverage results on Fashion MNIST. We present coverage percentage of the test dataset for three pre-determined risk-guarantees (one per column), where higher coverage is better, as well as negative log-likelihood and overall accuracy. The results indicate that SQUAD provides competitive uncertainty, especially at low-risk guarantees. Bayesian approximations via deep ensembles improve coverage all over the board, and for SQUAD in particular. When SGR can't guarantee the required risk level at high propability, 0\% coverage is reported. (\textit{2 std. deviations shown in parentheses, optimal results in \textbf{bold}.})}
\label{tab:cov}
\begin{center}
\begin{small}
\begin{tabular}{llllll}
\toprule
\textbf{Fashion MNIST} & cov@risk .5\% & cov@risk 1\% & cov@risk 2\% & NLL & Acc. \\
\midrule
Plain MLP & 29.1 ($\pm 20.71$) &  45.9 ($\pm 4.04$) & 60.4 ($\pm 3.17$) & 0.408 ($\pm .036$)  & 87.7 ($\pm .42$) \\
Maxout MCDropout & 41.9 ($\pm 9.86$) &  56.5 ($\pm 2.30$) &\textbf{ 69.9 }($\pm 1.48$) & 0.299 ($\pm .008$)  & \textbf{89.5} ($\pm .28$) \\
DLGM & 0.0 ($\pm .00$) &  33.5 ($\pm 2.42$) & 47.0 ($\pm 1.47$) & 0.446 ($\pm .007$)  & 84.3 ($\pm .15$) \\
SQUAD & \textbf{42.9} ($\pm 7.19$) &  \textbf{58.3} ($\pm 3.06$) & 69.5 ($\pm 1.55$) & \textbf{0.293} ($\pm .008$)  & \textbf{89.5} ($\pm .35$) \\
\midrule
\textbf{Deep Ensemble} & cov@risk .5\% & cov@risk 1\% & cov@risk 2\% & NLL & Acc.   \\
\midrule
Plain MLP Ensemble & 40.6  &  58.3 & 70.2  & 0.296 & 89.3 \\
Max. MCD. Ensemble & \textbf{48.2}  &  59.1 & 72.2  & \textbf{0.271} & \textbf{90.2} \\
DLGM Ensemble & 0.0  &  34.3 & 47.8  & 0.435 & 84.7 \\
SQUAD Ensemble & 47.5  & \textbf{61.6} & \textbf{73.1}  & 0.273 & 90.1 \\
\bottomrule
\end{tabular}
\end{small}
\end{center}
\end{table}

\begin{table}
\caption{SQUAD exhibits strong performance on a held-out dataset without hyperparameter tuning. }
\label{tab:cov_notmnist}
\begin{center}
\begin{small}
\begin{tabular}{llllll}
\toprule
\textbf{notMNIST} & cov@risk .5\% & cov@risk 1\% & cov@risk 2\% & NLL & Acc.  \\
\midrule
Plain MLP & 77.4 ($\pm 5.20$) &  85.5 ($\pm 1.91$) & 90.3 ($\pm .87$) & 0.228 ($\pm .009$)  & 93.3 ($\pm .29$)\\
Maxout MCDropout & 85.7 ($\pm 1.14$) &  90.6 ($\pm .62$) & 94.2 ($\pm .36$) & 0.165 ($\pm .003$)  & 95.3 ($\pm .21$)\\
SQUAD & \textbf{87.1} ($\pm 1.60$) &  \textbf{91.1} ($\pm .86$) & \textbf{94.5} ($\pm .50$) & \textbf{0.161} ($\pm .006$)  & \textbf{95.4} ($\pm .22$)\\
\midrule
Plain MLP Ensemble & 85.9  &  90.6 & 93.5  & 0.175 & 94.9 \\
Max. MCD. Ensemble & 88.5  &  92.8 & 95.7  & 0.148 & 96.0 \\
SQUAD Ensemble & \textbf{90.7}  &  \textbf{93.5 }& \textbf{96.1}  & \textbf{0.137} & \textbf{96.2} \\
\bottomrule
\end{tabular}
\end{small}
\end{center}

\caption{Results on SVHN indicate that the quantization scheme imposed by SQUAD models might hinder performance, but that this is effectively compensated by the SQUAD-factorized variant using a larger amount of bins. Even with $T=4$ MC samples at test time, SQUAD performs well.
}
\label{tab:cov_svhn}
\begin{center}
\begin{small}
\begin{tabular}{llllll}
\toprule
\textbf{MLP K=256} (SVHN) & cov@risk .5\% & cov@risk 1\% & cov@risk 2\% & NLL & Acc.  \\
\midrule
Plain MLP & 0.0 ($\pm .00$) &  0.0 ($\pm .00$) & 36.3 ($\pm 2.84$) & 0.758 ($\pm .065$)  & 83.1 ($\pm .42$)\\
Maxout MCDropout & 0.0 ($\pm .00$) &  50.7 ($\pm 1.42$) & 65.0 ($\pm 2.32$) & 0.480 ($\pm .020$)  & 86.4 ($\pm .71$)\\
SQUAD-factorized & \textbf{18.4} ($\pm 30.33$) &  \textbf{53.9} ($\pm 1.81$) & \textbf{66.7} ($\pm 2.10$) & \textbf{0.454} ($\pm .021$)  & \textbf{86.7} ($\pm .88$)\\
SQUAD & 1.7 ($\pm 6.75$) &  42.8 ($\pm 2.25$) & 59.3 ($\pm 1.38$) & 0.534 ($\pm .005$)  & 84.6 ($\pm .13$)\\
\midrule
Max. MCDropout T=4 & 0.0 ($\pm .00$) &  38.5 ($\pm 3.85$) & 57.6 ($\pm 2.66$) & 0.562 ($\pm .016$)  & 84.9 ($\pm .73$)\\
SQUAD-factorized T=4 & \textbf{10.7} ($\pm 26.25$) &  \textbf{49.9} ($\pm 3.51$) & \textbf{64.5} ($\pm 2.45$) & \textbf{0.480} ($\pm .024$)  & \textbf{86.2} ($\pm .67$)\\
SQUAD T=4 & 0.0 ($\pm .00$) &  38.0 ($\pm 1.86$) & 55.6 ($\pm .84$) & 0.569 ($\pm .019$)  & 83.7 ($\pm .52$)\\
\bottomrule
\end{tabular}
\end{small}
\end{center}
\end{table}
\subsection{Natural Images}
To explore larger models trained on natural image datasets, we lightly tune hyper-parameters on 256-latent 2-layer models over 100 TPE evaluations. As SVHN contains natural images in color, we anticipate a need for a higher amount of information per variable. We thus explore the effect of the matrix-factorized variant.

As shown in table~\ref{tab:cov_svhn}, SQUAD-factorized outperforms the non-factorized variant. Considering the computational cost at the optimum of a 4-neuron factorization ($B=4$) with $C$=37 quantization bins, the model clocks 3.4 million weights. In comparison, the optimum for the presented MCdropout results has $C$=11, using 9.0 million weights. On an NVIDIA Titan XP, the dropout baseline takes 13s per epoch on average, while SQUAD-factorized spans just 9s.

To evaluate the sample efficiency of the methods, we compare results at $T=4$ samples. We find that SQUAD's results suffer less from under-sampling than MCdropout. We tentatively attribute the sample efficiency to the flexible approximating posterior on the activations, which is in stark contrast to the rigid approximating distribution that MCdropout imposes on the weights. In conclusion, SQUAD comes out favorably in a resource-constrained environment.

\subsection{Analysis of latent variable distributions}
\label{sec:analysis}
\begin{figure}
\begin{center}
\centerline{\includegraphics[width=.9\columnwidth]{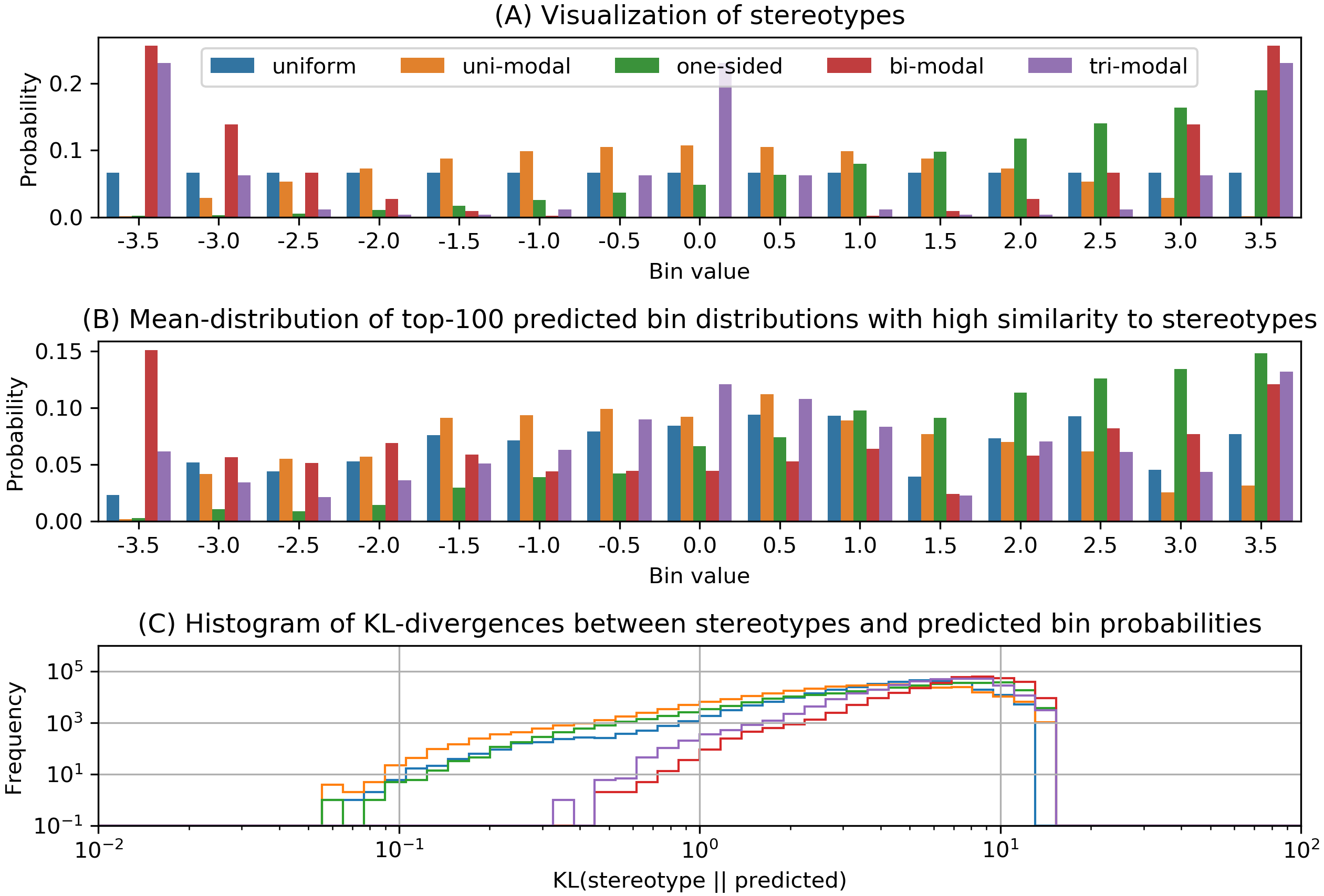}}
\caption{By analyzing the emerging predicted distributions of individual neurons in a converged SQUAD model, we find that the flexible variational distribution is used to its full advantage. Figure (A) visualizes a subset of interesting stereotypical distributions we hope to find in the model. Figure (B) summarizes distributions predicted by the model similar to stereotypes, discovered by looking at predicted distributions with low KL. Figure (C) shows how often distributions similar to stereotypes arise, as measured by the KL distance (lower KL is closer to stereotypes).}
\label{fig:stereotype1}
\end{center}
\end{figure}

In order to evaluate if the proposed variational distribution does not simply collapse into single mode predictions, we want to find out what type of distributions the model predicts over the latent variables. We visualize the forms of predicted distributions in figure~\ref{fig:stereotype1}a. Although this showcases only a small subset of potential multi-modal behavior that emerges, this demonstrates that the model indeed utilizes the distribution to its full potential. To provide an intuition on how these predicted distributions emerge, we present figure~\ref{fig:stereotype2} in the appendix.
\begin{figure}
\begin{center}
\centerline{\includegraphics[width=1.\columnwidth]{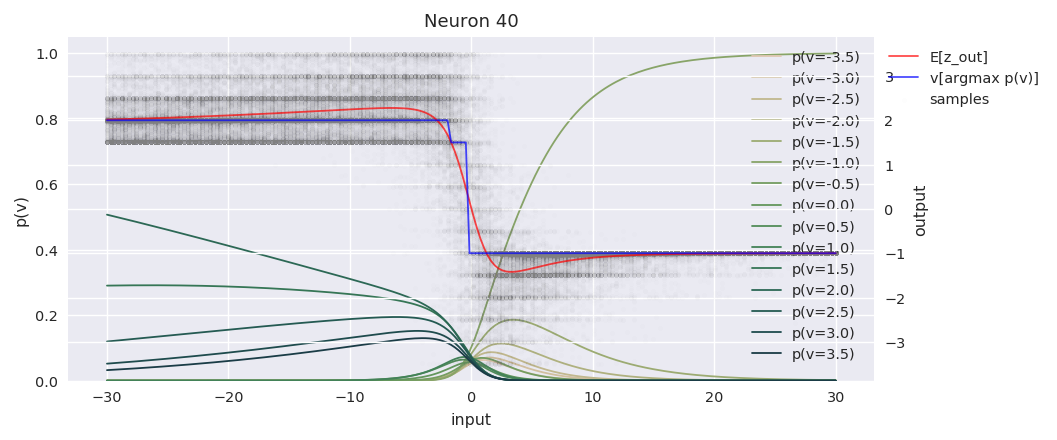}}
\caption{By using a 1-dimensional matrix factorization for a SQUAD-factorized distribution, we can visualize the type of (stochastic) activation functions learned by the method. After training the model as usual on fashion-MNIST, we take a random neuron from the first layer. We visualize how the predicted distribution of the output changes as a function of the 1-dimensional input. The left y-axis indicates the prob(ability per value as shown using the green line. The right y-axis indicates the value and in blue the most likely value is shown, and the gray dots represent samples from the neuron. The red line depicts the \textit{expected output} of the neuron. The shape of the expected output is akin to a peaky sigmoid activation, and similar shapes are found in the other neurons of the network as well. This provides food for thought on the design of activation functions for conventional neural networks. }
\label{fig:nonlinear}
\end{center}
\end{figure}

In figure~\ref{fig:nonlinear} we visualize one of the activation functions that the method learns for a 1-dimensional input SQUAD-factorized model. The learned activation functions resemble ``peaked'' sigmoid activations, which can be interpreted as a combination of an RBF kernel and sigmoid. This provides food for thought on how non-linearity's for conventional neural networks can be designed, and the effect of using such a non-linearity can be studied in further work.

\section{Discussion}
\label{sec:discuss}

In this work, we have proposed a new flexible class of variational distributions. To measure the effectiveness for real world classification, we applied the class to a deep variational information bottleneck model. By placing a quantization-based distribution on the activations, we can compute  uncertainty estimates over the outputs. We proposed an evaluation scheme motivated by the need in real-world domains to guarantee a minimal risk. The results presented indicate that SQUAD provides an improvement over plain neural networks and Gaussian information bottleneck models. In comparison to a MCDropout model, which approximates a Bayesian neural network, we get competitive performance. Moreover, qualitatively we find that the flexible distribution is used to its full advantage is sample efficient. The method learns interesting non-linearity's, is tractable and scaleable, and as the output domain is constrained, no batch normalization techniques are required. 

Various directions for future work arise. The improvement of ensemble methods over individual models indicates that there remains room for improvement for capturing the full uncertainty of the output, and thus a fully Bayesian approach to SQUAD which would include weight uncertainty, shows promise. The flexible class allows us to define a wide variety of interesting priors, which provides opportunity to study interesting priors that are hard to define as a continuous density. Likewise, more effective initialization of parameters for the proposed method requires further attention. Orthogonally, the proposed class can be applied to other variational objectives as well, such as the variational auto-encoder. Finally, the discretized nature of the variables allows for the analytical computation of other divergences such as mutual information and the Jensen-Shannon divergence, the effectiveness of which remains to be studied.

\paragraph{Acknowledgements}
We thank Bart Bakker, Maximilian Ilse, Dimitrios Mavroeidis, Jakub Tomczak, Daniel Worrall and anonymous reviewers for their insightful comments and discussions.
This research was supported by Philips Research, the SURFSara Lisa cluster and the NVIDIA GPU Grant. We thank contributors to TensorFlow (\cite{Abadi2016-vo}), Keras (\cite{chollet2015keras}) and Sacred (\cite{sacred}).

\bibliography{quantized_latent_paper}
\bibliographystyle{iclr2019_conference}

\section{Appendix}
\begin{figure*}
\begin{center}
\centerline{\includegraphics[width=\textwidth]{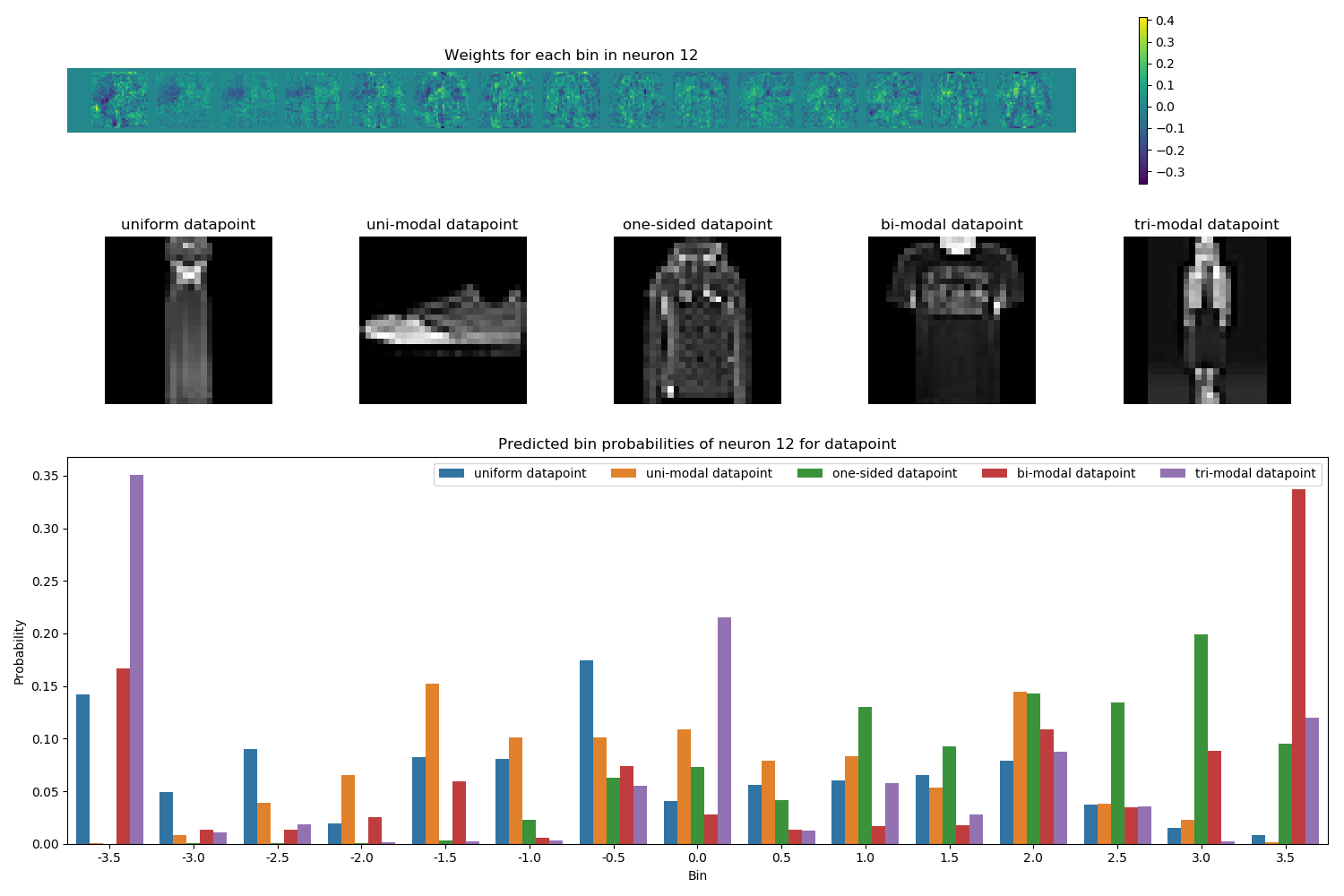}}
\caption{This figure serves to provide intuition on how a variety of distributions come about in our model. We show the set of weights used to predict the probability for the $C$ bins of a randomly selected latent variable $\zv_{l=1,k=12}$ from the first layer in a converged 2-layer SQUAD model (reshaped to a 28x28 squares for comparison with the data). We then present 5 data-points for which the neuron predicts a stereotypical distribution, as visualized in the last bar-plot.}
\label{fig:stereotype2}
\end{center}
\end{figure*}
\subsection{Effect of hyper-parameters on coverage:}
\begin{figure*}[h]
\begin{center}
\centerline{\includegraphics[width=\textwidth]{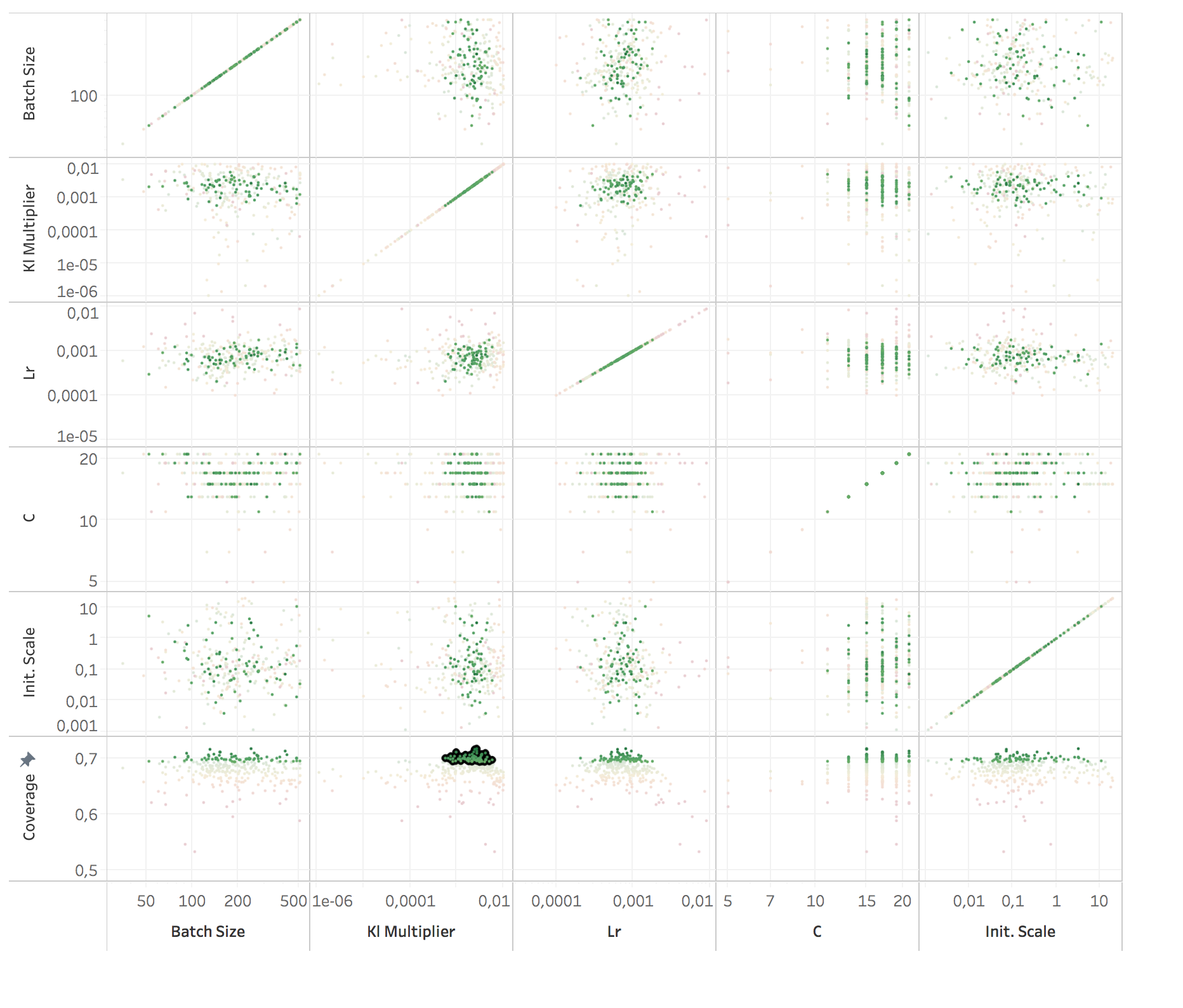}}
\caption{This figure visualizes the pairwise relationship between hyper-parameters of SQUAD and the effect on coverage. The top-60 configurations are highlighted. Green values are good, red values are bad. We have filtered on the optimal settings for bin values and prior to reduce clutter.}
\label{fig:hyperparams}
\end{center}
\end{figure*}

The optimal configuration of hyper-parameters and bin priors have been determined using 700 evaluations selected using TPE. The space of parameters explored is as follows, presented in the hyperopt API for transparency:
\begin{verbatim}
# Shared
C: quniform(2, 10, 1) * 2 + 1,
dropout rate: uniform(0.01, .95),
lr: loguniform(log(0.0001), log(0.01)),
batch_size: qloguniform(log(32), log(512), 1)
# SQUAD & Gaussian
kl_multiplier: loguniform(log(1e-6), log(0.01)),
init_scale: loguniform(log(1e-3), log(20)),
# SQUAD
use_bin_probs: choice(['uni', 'gaus']),
use_bins: choice(['equal_prob_gaus', 
                  'linearly_spaced']),
learn_bin_values: choice([
   'per_neuron', 'per_layer', 'fixed']),
\end{verbatim}
In figure~\ref{fig:hyperparams} we visualize the pairwise effect of these hyper-parameters on the coverage. The optimal configuration found in for the main SQUAD model are:
batch size: 244,
KL multiplier: 0.0027,
learn bin values: per layer,
$p(z)$: uniform,
$\vv$: linearly spread over (-3.5,3.5),
lr: 0.0008,
$C$: 15,
initialization scale: 3.214.

\end{document}

%% file: math_commands.tex
\usepackage{amsmath,amsfonts,bm}

\def\eqref#1{equation~\ref{#1}}

\def\1{\bm{1}}

\def\vv{{\bm{v}}}

\DeclareMathAlphabet{\mathsfit}{\encodingdefault}{\sfdefault}{m}{sl}
\SetMathAlphabet{\mathsfit}{bold}{\encodingdefault}{\sfdefault}{bx}{n}

\newcommand{\E}{\mathbb{E}}

\newcommand{\KL}{D_{\mathrm{KL}}}

\DeclareMathOperator*{\argmax}{arg\,max}